\documentclass[a4paper]{article}

\usepackage{INTERSPEECH2015}

\usepackage{amssymb,bm}
\usepackage{textcomp}

\usepackage{amsmath,dsfont,graphicx}
\usepackage{mathrsfs}
\usepackage{epstopdf}

\usepackage{tikz}
\usetikzlibrary{decorations.pathmorphing} 
\usetikzlibrary{fit}					  
\usetikzlibrary{backgrounds}	          
\usetikzlibrary{shapes}
\usetikzlibrary{positioning}

\newcommand{\ignore}[2]{\hspace{0in}#2}

\ninept

\title{Shared latent subspace modelling within Gaussian-Binary Restricted Boltzmann Machines for NIST i-Vector Challenge 2014}

\makeatletter
\def\name#1{\gdef\@name{#1\\}}
\makeatother
\name{{\em Danila Doroshin$^1$, Alexander Yamshinin$^1$, Nikolay Lubimov$^1$,}\\
      {\em Marina Nastasenko$^1$, Mikhail Kotov$^1$, Maxim Tkachenko$^1$}}
\address{$^1$Stel - Computer Systems Ltd., Moscow, Russia \\
  {\small \tt \{doroshin, yamshinin, lubimov, marina.nastasenko, kotov, tkachenko\}@stel.ru}
}

\begin{document}

\maketitle
\begin{abstract}
This paper presents a novel approach to speaker subspace modelling based on Gaussian-Binary Restricted Boltzmann Machines (GRBM). The proposed model is based on the idea of shared factors as in the Probabilistic Linear Discriminant Analysis (PLDA). GRBM hidden layer is divided into speaker and channel factors, herein the speaker factor is shared over all vectors of the speaker. Then Maximum Likelihood Parameter Estimation (MLE) for proposed model is introduced. Various new scoring techniques for speaker verification using GRBM are proposed. The results for NIST i-vector Challenge 2014 dataset are presented.
\end{abstract}

\section{Introduction}

Actual approaches to text-independent automatic speaker verification (ASV) generally focus on the modelling of speaker and channel variability.  The background of majority of these methods is based on factorising of the long-term distribution of spectral features. The standard method in ASV is to model this distribution using Gaussian Mixture Model (GMM) which is trained on a large audio database and referred as Universal Background Model (UBM). The Joint Factor Analysis technique \cite{kenny2005joint} is based on decomposition of a UBM supervector\footnote[1]{A supervector is a vector of stacked GMM mean vectors} into the additive components belonging to speaker and channel subspace. Speaker and channel subspaces are modeled using low-dimensional factors. The i-vector approach is based on the total variability model \cite{dehak2011front} representing supervector in the low-dimensional space containing both speaker and channel information. Probabilistic Linear Discriminant Analysis (PLDA) \cite{prince2007probabilistic} is applied to handle the influence of the channel variability in the i-vector space. PLDA deals with the decomposition of i-vectors on speaker and channel factors where the speaker factor is the same for all i-vectors of the speaker \cite{kenny2008study}.

In this paper we examine an alternative way to effectively model speaker subspace using Restricted Boltzmann Machines (RBM). The idea is close to the PLDA factor modelling and based on dividing RBM hidden layer into the speaker and channel factors where the speaker factor is shared over all vectors of the speaker. The proposed model uses Gaussian-Binary RBM (GRBM) in contrast to a model described in \cite{stafylakis2012plda} where Gaussian-Gaussian RBM was considered. The proposed approach is simply extended to the case of Binary-Binary RBM (BRBM).
This choice is motivated by the ability of using Gaussian-Binary and Binary-Binary blocks as the internal parts of deeper architectures as Deep Belief Networks and Deep Boltzmann Machines.

The paper is organized as follows. In section \ref{sec:GBRBM}, the basic definitions of GRBMs are covered, then GRBM with shared latent subspace and corresponding generative model is introduced, MLE for proposed model including modification of contrastive divergence algorithm is performed. In section \ref{sec:Scoring}, various new scoring techniques for ASV are described including log-likelihood ratio (LLR) and normalized cosine scoring. In section \ref{sec:experiments} the train and test datasets are described. The results for NIST i-vector Challenge 2014 dataset are given and compared to the baseline and state-of-the-art methods. In section \ref{sec:conclusions}, conclusions and future work directions are discussed. In the appendix section \ref{sec:appendix} some theoretical proofs are presented.

\section{Shared latent subspace modelling within Gaussian-Binary Restricted Boltzmann Machines}
\label{sec:GBRBM}

\subsection{General GRBM}
\label{ssec:general}
GRBM defines probability density function (PDF) with input visible variable $x$ and hidden (latent) variable $h$ \cite{salakhutdinov2009learning, hinton2010practical}
\begin{equation*}
  P(x,h) = \frac{1}{Z}e^{-E(x,h)}
\end{equation*}
where $Z$ is a normalizing constant called a partition function and $E(x,h)$ is an energy function.
For GRBM $x$ is from the continuous space $\mathds{R}^p$ and $h$ is from the discrete space $h \in \{0,1\}^r$.
$E(x,h)$ depends on visible bias $b$, hidden bias $d$,
vector of standard deviations $\sigma$ and connectivity matrix $W$
\begin{equation*}
  E(x,h) = \frac{1}{2}\|\frac{x-b}{\sigma}\|^2 - d^Th - \left(\frac{x}{{\sigma^2}}\right)^TWh
\end{equation*}
Here and then $\star/\star$ denotes element-wise division of vectors, $\star^2$ denotes element-wise squaring, $T$ denotes transposition and $\|\star\|$ is Euclidean norm.
\subsection{GRBM with shared latent subspace}
\label{ssec:fGRBM}
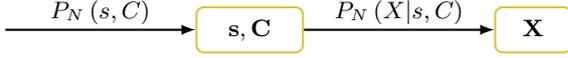
\begin{figure}[htbp]
\centering
\tikzstyle{visible}=[rounded corners=1mm,
                                    thick,
                                    minimum height=0.6cm,
                                    minimum width=1.0cm,
                                    draw=yellow!80!black,
                                    fill=white]

\tikzstyle{latent}=[rounded corners=1mm,
                                    thick,
                                    minimum height=0.6cm,
                                    minimum width=1.4cm,
                                    draw=yellow!80!black,
                                    fill=white]

\tikzset{dotstyle/.style={->}}
\tikzset{standard/.style={->,sloped,anchor=south,auto=false}}

\begin{tikzpicture}[>=latex,text height=1.5ex,text depth=0.25ex, node distance=1.4ex]
  
  
        \node (NO)    {};
        \node[latent]  (H) [right =2.5cm of NO]   {$\mathbf{s},\mathbf{C}$};
        \node[visible] (X) [right =2.5cm of H]  {$\mathbf{X}$};

    \path
        (NO)   edge[standard,thick] node[align=center] {$P_N\left(s,C\right)$}      (H)
        (H)   edge[standard,thick] node[align=center] {$P_N\left(X|s,C\right)$}     (X)
        ;
\end{tikzpicture}

\caption{The generative process for $N$-order model}
\label{fig:generative}
\end{figure}

GRBM is modified to simulate speaker and channel variability. The hidden variable is divided into the speaker factor $s$
and the channel factor $c$, i.e. $h = [s; c]$. According to this, parameters are split into two groups, i.e. $d = [f; g]$, $W = [F, G]$. Rewrite the energy function expression using split parameters
\begin{equation*}
  E(x,s,c) = \frac{1}{2}\|\frac{x-b}{\sigma}\|^2 - f^Ts - g^Tc - \left(\frac{x}{\sigma^2}\right)^T (F s + G c)
\end{equation*}
The speaker factor is supposed to
be the same for all i-vectors of one speaker while the channel factors are
individual for each i-vector. Below a set of generative models depending on the number of i-vectors corresponding to the speaker and PDF for them are introduced. Consider the case of $N$ i-vectors of the speaker and correspondent $N$-order generative model.
Denote speaker data as $X=\{x_1, x_2, \dots, x_N\}$, channel factors as $C=\{c_1, c_2, \dots, c_N\}$ and speaker factor as $s$. PDF for $N$-order model is expressed as follows
\begin{equation}
  \label{SLSRBM_pdf}
  P_N(X,s,C) = \frac{1}{Z_N}e^{-E_N(X,s,C)}
\end{equation}
where $E_N(X,s,C) = \sum_{n=1}^{N} E(x_n,s,c_n)$ and
$Z_N = \int_X \sum_{s, C}{e^{-E_N(X,s,C)}} dX$.
The generative process for this model is shown in Figure~\ref{fig:generative}. 
First $s, C$ are generating according the distribution $P_N(s,C) = \int P_N(X,s,C) dX$ then $X$ is generating according the distribution $P_N(X |s,C)$, where
\begin{equation}
    \label{SLSRBM_normalPDF}
    P_N(X |s,C) = \prod_{n=1}^{N} \mathcal{N}(x_n, b + Fs + Gc_n, \sigma^2)
\end{equation}
and $\mathcal{N}$ denotes Gaussian distribution.
\subsection{Maximum likelihood parameter estimation}
\label{ssec:ML}
Assume we have a labeled training set of $K$ speakers, denoted by $\mathds{X}=\{X_k\}_{k=1}^{K}$
where $X_k$ is data with $N_k$ i-vectors that corresponds to $k$-th speaker. Let $N_k \in \{2, 3, \cdots, M\}$, hence there are $M-1$ generative models, and it is assumed that their parameters are tied. The aim is to estimate the set of parameters $\Theta = \{f, g, F, G, b, \sigma\}$ using MLE criterium that is standard approach for RBMs \cite{hinton2010practical}. For the optimization of MLE objective function we use a stochastic gradient descent approach that is widely used for RBMs \cite{wang2014gaussian, salakhutdinov2009learning}. Since data for each pair of speakers are assumed to be independent,
normalized log-likelihood function takes the form of sum of log-likelihood functions for each generative model
\begin{eqnarray*}
  \mathcal{L}_{norm}(\mathds{X}|\Theta) =
  \frac{1}{\sum_{k} {N_k}} \sum_{k} \mathcal{L}(X_k|\Theta) = \\
  =\frac{1}{\sum_{k} {N_k}} \sum_{N = 2}^{M} \sum_{k: N_k = N} \mathcal{L}(X_k|\Theta)\nonumber
\end{eqnarray*}
Further speaker's index will be neglected and there will be discussed the
likelihood of the data from one speaker. Denote speaker data as $X=\{x_1, x_2, \dots, x_N\}$ then
%
\begin{equation}
  \label{RBM_LL_one}
  \mathcal{L}(X|\Theta) = \log{P_N(X)}
\end{equation}
%
Denote the realization of speaker factor as $s$ and the realizations of channel factors as $c_n$, $C = \{c_1, c_2, \dots, c_N\}$. Consider log-likelihood from \eqref{RBM_LL_one} marginalizing \eqref{SLSRBM_pdf} over all possible values of latent variables
\begin{equation}
  \label{RBM_constr}
  \mathcal{L}(X|\Theta) = \log{\sum_{s, C} e^{-E_N(X,s,C)}} - \log{Z_N}
\end{equation}
Consider the first part of gradient of \eqref{RBM_constr}, making the same transformations as for the general GRBM \cite{hinton2010practical}
\begin{eqnarray*}
\nabla_{\Theta}{\mathcal{L}_1(X|\Theta)} \ignore{ - \frac{\sum_{s, C} e^{-E_N(X,s,C)}\frac{\partial}{\partial \Theta}E_N(X,s,C)}{\sum_{s, C} e^{-E_N(X,s,C)}} = \\}
 = - \frac{\sum_{s, C} P_N(X,s,C)\frac{\partial}{\partial \Theta}E_N(X,s,C)}{P_N(X)} = \\
 = - \sum_s \sum_{n=1}^{N} \sum_{c_n} P_N(s, c_n| X) \frac{\partial E(x_n, s, c_n)}{\partial \Theta}
\end{eqnarray*}
As a result, the gradient of \eqref{RBM_constr} is represented as the following sum
\begin{equation}
\label{RBM_logPDF_diff}
\nabla_{\Theta}{\mathcal{L}(X|\Theta)} = \nabla_{\Theta}{\mathcal{L}_1(X|\Theta)} - \mathcal{E}_{P_N(\tilde{X})} \left[ \nabla_{\Theta}{\mathcal{L}_1(\tilde{X}|\Theta)} \right]
\end{equation}
Here $\mathcal{E}$ denotes expectation: $\mathcal{E}_{P_N(\tilde{X})} \left[ \star \right] = \int_{\tilde{X}} P_N(\tilde{X}) \star d\tilde{X}$.
The modification of the contrastive divergence algorithm \cite{hinton2010practical} that enables to compute the second term of the gradient \eqref{RBM_logPDF_diff} is presented in section \ref{sec:CV}. Below the gradient of the first term will be considered. 
Taking into account the derivatives of energy function \cite{wang2014gaussian}
the gradient of $\mathcal{L}_1(X|\Theta)$ takes the following form
\begin{eqnarray}
\label{F_diff}
&&\nabla_{F_{ij}}\mathcal{L}_1(X|\Theta) =
P_N(s_j=1|X)\frac{\bar{x}_i}{{\sigma_i}^2}\\
\label{d_diff}
&&\nabla_{f_i}{\mathcal{L}_1(X|\Theta)} =
N P_N(s_j=1|X) \ignore{- N \mathds{E}_{P(x)} \left[P(s(j)=1|x)\right]} \\
\label{G_diff}
&&\nabla_{G_{ij}}{\mathcal{L}_1(X|\Theta)} =
\sum_{n} P_N(c_{nj}=1|X)\frac{ x_{ni}}{{\sigma_i}^2}\\
\label{q_diff}
&&\nabla_{g_i}{\mathcal{L}_1(X|\Theta)} =
\sum_{n} P_N({c_{nj}}=1|X)\\
\label{b_diff}
&&\nabla_{b_i}{\mathcal{L}(X|\Theta)} =
\frac{1}{\sigma_i^2}\left(\bar{x}_i - N b_i \right)\\
\label{logsigma_diff}
&&\nabla_{z_i}{\mathcal{L}_1(X|\Theta)} =
-\frac{\bar{x}_i}{\sigma_i^2} \sum_j F_{ij} P_N(s_j=1|X) -\\ 
&&-\sum_{n,j}{ \frac{x_{ni}}{\sigma_i^2} G_{ij}P_N(c_{nj}=1|X)} + \frac{1}{2} \sum_n{\frac{(x_{ni}-b_i)^2}{\sigma_i^2}}  \nonumber
\end{eqnarray}
Here and further $\bar{x} = \sum_{n=1}^{N}{x_n}$ and $i,j$ denote indexing over dimensions. Additionally, instead of $\sigma_i$, we update log-variances $z_i = \log{\sigma_i^2}$ which are naturally constrained to stay positive \cite{cho2011improved}.
Posteriori probabilities of latent factors from the expressions (\ref{F_diff}-\ref{logsigma_diff}) are determined from the following relations, which are proved
in the appendix section of the paper
\begin{eqnarray}
\label{sp_factor_poster}
P_N(s_j=1|X) = sigm\left(N f_j + \left({\bar{x}}/{\sigma^2}\right)^T F_{*j}\right)\\
\label{ch_factor_poster}
P_N(c_{nj}=1|X) = sigm\left(g_j + \left({x_n}/{\sigma^2}\right)^T G_{*j}\right)
\end{eqnarray}
Here and further $F_{*j}$ and $G_{*j}$ denotes respectively $j$-th column of the matrices and $sigm(\star) = 1/\left(1 + e^{-\star}\right)$. From the expression \eqref{ch_factor_poster} it is clear that posterior for $c_n$ depends only on $x_n$ and it is the same as for the general GRBM. The main difference is that all speaker's i-vectors $X$ influence the speaker factor posterior \eqref{sp_factor_poster}.
\subsubsection{Contrastive divergence}
\label{sec:CV}
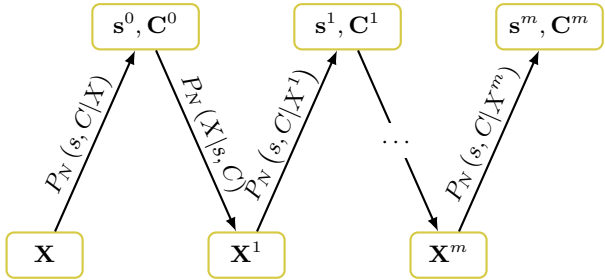
\begin{figure}[htbp]
\centering
\tikzstyle{visible}=[rounded corners=1mm,
                                    thick,
                                    minimum height=0.6cm,
                                    minimum width=1.0cm,
                                    draw=yellow!80!black,
                                    fill=white]

\tikzstyle{latent}=[rounded corners=1mm,
                                    thick,
                                    minimum height=0.6cm,
                                    minimum width=1.4cm,
                                    draw=yellow!80!black,
                                    fill=white]

\tikzset{dotstyle/.style={->}}
\tikzset{standard/.style={->,sloped,anchor=south,auto=false}}

\begin{tikzpicture}[>=latex,text height=1.5ex,text depth=0.25ex, node distance=1.4ex]
  
  
        \node[visible] (X)    {$\mathbf{X}$};
        \node[latent]  (H)   [above right =2.4cm and 0.1cm of X]   {$\mathbf{s}^0,\mathbf{C}^0$};
        \node[visible] (X^1) [below right =2.4cm and 0.1cm of H]  {$\mathbf{X}^1$};
        \node[latent]  (H^1) [above right =2.4cm and 0.1cm of X^1] {$\mathbf{s}^1,\mathbf{C}^1$};
        \node[visible] (X^m) [below right =2.4cm and 0.1cm of H^1]  {$\mathbf{X}^m$}; 
        \node[latent]  (H^m) [above right =2.4cm and 0.1cm of X^m] {$\mathbf{s}^m,\mathbf{C}^m$};

    \path
        (X)   edge[standard,thick] node[align=center] {$P_N\left(s,C|X\right)$}    (H)
        (H)   edge[standard,thick] node[align=center] {$P_N\left(X|s,C\right)$}    (X^1)
        (X^1) edge[standard,thick] node[align=center] {$P_N\left(s,C|X^1\right)$}  (H^1)
        (H^1) edge[dotstyle,thick] node[fill=white]   {$\cdots$} (X^m)
        (X^m) edge[standard,thick] node[align=center] {$P_N\left(s,C|X^m\right)$}  (H^m)
        ;
\end{tikzpicture}

\caption{Contrastive divergence}
\label{fig:CD}
\end{figure}

The modification of the contrastive divergence algorithm \cite{hinton2010practical} is presented below. It enables to compute approximately the second part of the gradient \eqref{RBM_logPDF_diff}. Expectation is replaced by mean over a finite set of samples from distribution $P_N(\tilde{X})$. Since it is hard to get these samples because of the complexity of the generative process, 
an approximate algorithm called the m-steps contrastive divergence \cite{hinton2002training} is applied. Algorithm scheme is presented in Figure~\ref{fig:CD}.
Data of the speaker $X$ is used to initialize the algorithm on the zero step. Intermediate k-th step of the algorithm is presented below.
Reconstruction of visible data $X^k$ is sampled using \eqref{SLSRBM_normalPDF}. Latent variables $s^k, C^k$ are sampled using \eqref{sp_factor_poster} and \eqref{ch_factor_poster}. For binarization we use uniformly distributed random thresholds following recommendations from \cite{hinton2010practical}.
%
%
%
\subsection{Scoring}
\label{sec:Scoring}
In this section various scoring strategies for GRBM with shared latent subspace will be presented.
\subsubsection{Log-likelihood scoring}
\label{sec:LLR_scoring}
The LLR for a given verification trial $\{X, x_t\}$, i.e. $X$ is a set of $N$ enrollment speaker's vectors and $x_t$ is a test vector, 
is the LLR between target and non-target hypotheses.The target
hypothesis is that the trial vectors share a common speaker factor, i.e. generated by $N+1$-order model. Non-target hypothesis is that $X$ is generated by $N$-order model and $x_t$ is independent from them and generated by $1$-order model.
\begin{equation*}
l = \log{ \frac{P_{N+1}(X,x_t)}{P_N(X) P_1(x_t)} }
\end{equation*}
The expression for the LLR score is given below and its proof is given in the appendix section
\begin{eqnarray*}
l = \sum_i \log{ \frac{1 + e^{(N+1) f_i + \left({\left(\bar{x} + x_t\right)}/{\sigma^2}\right)^T F_{*i}}}
{{\left(1 + e^{N f_i + \left({\bar{x}}/{\sigma^2}\right)^T F_{*i}}\right)}
{\left(1 + e^{f_i + \left({x_t}/{\sigma^2}\right)^T F_{*i}}\right)}} } + \\ 
+ \log{\frac{Z_{N}Z_{1}}{Z_{N+1}}}
\end{eqnarray*}
Some methods exist for the approximate computation of the partition function \cite{burda2014accurate}. Note that values of the partition function do not influence the performance of the system in case when all speakers have the same number of enrollment vectors.
\subsubsection{Cosine scoring}
\label{sec:cos_scoring}
We apply the standard cosine scoring \cite{senoussaoui2010vector} to i-vectors previously projected onto the subspace $F^T$.
Denote $y_n=\frac{F^Tx_n}{\|F^Tx_n\|}$ for each speaker's i-vector from $X$ and $y_t=\frac{F^Tx_t}{\|F^Tx_t\|}$ for test.
The score is cosine between average speaker's vector $y_{sp}={\sum_n y_n}/N$ and the test vector
\begin{equation*}
l_{cos} = y_t^T\frac{y_{sp}}{\|y_{sp}\|}
\end{equation*}
In addition to the general cosine score, we propose normalized cosine score $l_{norm}$. It takes into account information on the width of the speaker's cluster that is lost in the standard cosine scoring.
General cosine score is divided by the
average cosine within the speaker's set $cos_{sp} = \sum_n y_n^T\frac{y_{sp}}{\|y_{sp}\|}/N$.
It can be shown that $cos_{sp} = \|y_{sp}\|$.
Taking it into account, the expression for the normalized cosine score takes the form
\begin{equation*}
l_{norm} = \frac{l_{cos}}{\|y_{sp}\|}
\end{equation*}
\subsubsection{PLDA on F-projected i-vectors}
\label{sec:plda_scoring}
PLDA model is trained on i-vectors projected onto the subspace $F^T$ and then projected on unit sphere -- $y_n$. PLDA handles residual channel variability using linear factor model \cite{prince2007probabilistic}. Scoring is done using the LLR for PLDA model \cite{larcher2013phonetically, rajan2014single}.
\section{Experimental results}
\label{sec:experiments}
\subsection{Dataset}
\label{ssec:dataset}
\textit{NIST i-vector Machine Learning Challenge 2014} dataset has been chosen to test the efficiency of the proposed model. The dataset consists of a labeled development set (\textit{devset}), a labeled model set (\textit{modelset}) with 5 i-vectors per model and an unlabeled test set (\textit{testset}). Since labels for the \textit{devset} were not available during the challenge, the best results were obtained from methods that allowed to cluster the  \textit{devset} and then to apply PLDA \cite{khoury2014hierarchical, novoselov2014stc}.

In our experiments we reformed the dataset. Preliminary all i-vectors with duration less then 10 seconds have been removed for their bad quality \cite{khoury2014hierarchical, novoselov2014stc}. We construct a new labeled \textit{trainset}, \textit{modelset}, \textit{testset}, \textit{modelsetCV}, \textit{testsetCV}.
Speakers from \textit{devset} with 3 to 10 i-vectors united with the initial \textit{modelset} are assigned to the \textit{trainset}, with 11 to 15 i-vectors are assigned to the new \textit{modelset} and \textit{testset}, remaining speakers with more then 15 i-vectors form cross validation set (\textit{modelsetCV}, \textit{testsetCV}). First 5 i-vectors from each speaker's set form enrollment in the \textit{modelset} and the remaining form the \textit{testset}. The same is done for the cross validation set. Eventually the \textit{trainset} contains 3281 speakers and total 18759 i-vectors, 717 speakers with 3585 i-vectors and  5400 i-vectors in the \textit{modelset} and the \textit{testset} respectively. We used minDCF as a measure of the system performance and a measure for the cross validation processing
\begin{equation*}
minDCF = \min_{th} {FR}(th) + 100 {FA}(th)
\end{equation*}
where $FA$ and $FR$ denote the false acceptance and the false rejection rates,
and $th$ the varying threshold. The trials
consist of all possible pairs involving a target speaker set from the \textit{modelset} and a test
i-vector from the \textit{testset}.
\subsection{Parameters estimation}
\label{ssec:params}
Whitened \cite{garcia2011analysis} \textit{trainset} is used for the parameter estimation. The parameters of whitening are computed on the \textit{trainset} too. This transform is used further for all trials.
We set initial biases $f$, $g$ and $b$ to the zero. Following the recommendations from \cite{hinton2010practical} elements of the connectivity matrices $F$ and $G$ are generated using normal distribution with zero mean and standard deviation equal to 0.01. Elements of standard deviation vector $\sigma$ were set to 1.0. The case of $\sigma$ reestimation showed the worse results.

The best performance was obtained using the speaker factor dimension equal to 500, the channel factor dimension equal to 100 while i-vector dimension equal to 600.
We used the mini-batch stochastic gradient descent algorithm \cite{hinton2010practical} with learning rate 0.01, momentum 0.5 and zero weight decay. Each batch contained 256 speakers. After each epoch the speakers are shuffled between batches. It took 40 epochs to achieve the best minDCF on the cross validation set. In case when all speakers belong to one batch it took 10 times more iterations to reach the same performance of the system. 
To train PLDA model on i-vectors, whitened \textit{trainset} was projected on the unit sphere \cite{garcia2011analysis}. It was found that the best speaker and channel factor dimensions for PLDA are equal to 590 and 10 respectively. PLDA model trained on i-vectors that were projected on $F^T$ has the speaker and channel factor dimensions equal to 499 and 1 respectively. Increase of the channel factor dimension showed the worse results.
\subsection{Results}
\label{ssec:results}
We compare our algorithm with the NIST 2014 baseline cosine scoring and the state of the art \cite{khoury2014hierarchical, novoselov2014stc} PLDA.
\begin{figure}[htb]

\begin{minipage}[b]{1.0\linewidth}
  \centering
  \centerline{\includegraphics[width=12.5cm]{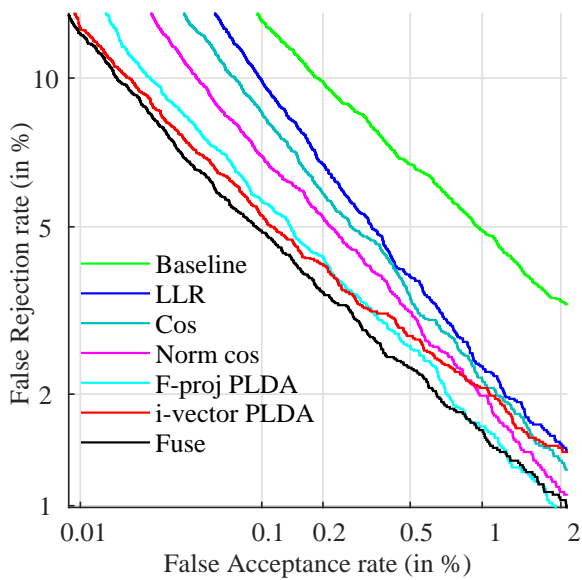}}
\end{minipage}

\caption{Comparison of the proposed scoring algorithms
  with NIST 2014 baseline cosine and PLDA }\medskip
\label{fig:DET}
\end{figure}
\begin{table}[h]
    \begin{center}
        \begin{tabular}{  c  c  c  c  c  c  c }
        \hline
                    & Baseline  & LLR   & Cos\\ \hline
        EER (in \%) & 2.81      & 1.68  & 1.58\\
        minDCF      & 0.210     & 0.185 & 0.167\\ 
        \\ \hline
        Norm cos    & F-proj PLDA   & i-vector PLDA     & Fuse\\ \hline
        1.43        & \textbf{1.30} & 1.51              & 1.33\\ 
        0.145       & 0.123         & \textbf{0.114}    & \textit{\textbf{0.108}}\\
        \end{tabular}
        \caption{Results on NIST 2014 dataset.}
        \label{table1}
    \end{center}
\end{table}
\noindent
As the results in Table \ref{table1} and Figure~\ref{fig:DET} demonstrate,
all scoring strategies perform better then challenge baseline.
Despite the optimality of log-likelihood GRBM scoring, it did not show
the best results among the other GRBM scoring strategies.
Perhaps, this is due to the specific of the i-vector data.
The considered normalized cosine scoring performs better then the standard cosine scoring.
In terms of EER, the best result is achieved on PLDA trained on i-vectors projected on speaker space $F^T$ of GRBM.
In addition, linear fuse \cite{brummer2007fusion} of two PLDA models is presented. The first model uses i-vectors as features and the second one uses i-vectors projected on $F^T$. Coefficients of the fuse were estimated on the cross validation set
by using logistic regression training  with weighted MLE criterium \cite{van2013distribution}. As can be seen in Figure~\ref{fig:DET}, the fused scores outperform i-vector PLDA in the area of low FA and retain performance in the EER area.
\section{Conclusions and Further Work}
\label{sec:conclusions}
We used shared latent subspace in GRBM hidden layer to separate speaker dependent and speaker independent factors in i-vector space. Approximate maximum likelihood parameters estimation is presented. For the proposed model several scoring methods for the speaker verification were considered , including a novel log-likelihood scoring and normalized cosine scoring. PLDA operating with i-vectors projected on GRBM speaker space performed results that are comparable to the state of the art i-vector PLDA approach. Fuse of these two PLDA models showed the best results at all operating points.

In further work, the method of projection on GRBM speaker space can be viewed as a stand-alone channel variability compensation technique. GRBM with shared latent subspace can be extended to the other types of RBM and can be used as a block in deeper architectures. 
\section{Appendix}
\label{sec:appendix}
In this section proofs of LLR score expression from section \ref{sec:Scoring} and expressions \eqref{sp_factor_poster}, \eqref{ch_factor_poster} are derived. They can be obtained if there is an
expression for a posterior probability $P_N(s, C|X)$.
First we derive joint PDF for latent variables and data $P_N(s, C, X)$
using its definition \eqref{SLSRBM_pdf}
\begin{eqnarray}
\label{appendix1}
&&P_N(s, C, X) = \mathcal{C}_{N,X} \cdot\\
&&\prod_{i,n,j} e^{N f_i s_i + \left({\bar{x}}/{\sigma^2}\right)^T F_{*i} s_i}
e^{g_j c_{nj} + \left({x_n}/{\sigma^2}\right)^T G_{*j}c_{nj}} \nonumber
\end{eqnarray}
Here $\mathcal{C}_{N,X} = \frac{1}{Z_N} e^{-\frac{1}{2}\sum_n \|\frac{x_n - b}{\sigma}\|^2}$.
Marginalizing \eqref{appendix1} over
all possible latent variables we have
\begin{eqnarray}
\label{appendix2}
&&P_N(X) = \mathcal{C}_{N,X} \cdot\\
&&\prod_{i,n,j}\left(1 + e^{N f_i + \left({\bar{x}}/{\sigma^2}\right)^T F_{*i}}\right)
\left(1 + e^{g_j + \left({x_n}/{\sigma^2}\right)^T G_{*j}}\right) \nonumber
\end{eqnarray}
Eventually the posterior probability of latent variables is the division of \eqref{appendix1}
on \eqref{appendix2}
\begin{eqnarray}
\label{appendix3}
&&P_N(s, C|X) =\\
&&\prod_{i} \frac{ e^{N f_i s_i + \left({\bar{x}}/{\sigma^2}\right)^T F_{*i} s_i} }
{ 1 + e^{N f_i + \left({\bar{x}}/{\sigma^2}\right)^T F_{*i}} }
\prod_{n,j} \frac{e^{g_j c_{nj} + \left({x_n}/{\sigma^2}\right)^T G_{*j} c_{nj}}}
{1 + e^{g_j + \left({x_n}/{\sigma^2}\right)^T G_{*j}}} \nonumber
\end{eqnarray}
Now expressions for \eqref{sp_factor_poster}, \eqref{ch_factor_poster} can be obtained by summing
\eqref{appendix3} over corresponding latent variables. The expression for LLR score is obtained by applying \eqref{appendix2} to the three trial subsets.
\section{Acknowledgement}
\label{sec:acknowledgement}
Research is conducted by Stel - Computer systems ltd. with support of the Ministry of Education and Science of the Russian Federation (Contract №14.579.21.0058)
Unique ID for Applied Scientific Research (project) RFMEFI57914X0058.
The data presented, the statements made, and the views expressed are solely the responsibility of the authors.

\newpage
\eightpt

\bibliographystyle{IEEEtran}
\bibliography{RBMinter}

\begin{thebibliography}{10}
\providecommand{\url}[1]{#1}
\csname url@samestyle\endcsname
\providecommand{\newblock}{\relax}
\providecommand{\bibinfo}[2]{#2}
\providecommand{\BIBentrySTDinterwordspacing}{\spaceskip=0pt\relax}
\providecommand{\BIBentryALTinterwordstretchfactor}{4}
\providecommand{\BIBentryALTinterwordspacing}{\spaceskip=\fontdimen2\font plus
\BIBentryALTinterwordstretchfactor\fontdimen3\font minus
  \fontdimen4\font\relax}
\providecommand{\BIBforeignlanguage}[2]{{%
\expandafter\ifx\csname l@#1\endcsname\relax
\typeout{** WARNING: IEEEtran.bst: No hyphenation pattern has been}%
\typeout{** loaded for the language `#1'. Using the pattern for}%
\typeout{** the default language instead.}%
\else
\language=\csname l@#1\endcsname
\fi
#2}}
\providecommand{\BIBdecl}{\relax}
\BIBdecl

\bibitem{kenny2005joint}
P.~Kenny, ``Joint factor analysis of speaker and session variability: Theory
  and algorithms,'' \emph{CRIM, Montreal,(Report) CRIM-06/08-13}, 2005.

\bibitem{dehak2011front}
N.~Dehak, P.~Kenny, R.~Dehak, P.~Dumouchel, and P.~Ouellet, ``Front-end factor
  analysis for speaker verification,'' \emph{Audio, Speech, and Language
  Processing, IEEE Transactions on}, vol.~19, no.~4, pp. 788--798, 2011.

\bibitem{prince2007probabilistic}
S.~J. Prince and J.~H. Elder, ``Probabilistic linear discriminant analysis for
  inferences about identity,'' in \emph{Computer Vision, 2007. ICCV 2007. IEEE
  11th International Conference on}.\hskip 1em plus 0.5em minus 0.4em\relax
  IEEE, 2007, pp. 1--8.

\bibitem{kenny2008study}
P.~Kenny, P.~Ouellet, N.~Dehak, V.~Gupta, and P.~Dumouchel, ``A study of
  interspeaker variability in speaker verification,'' \emph{Audio, Speech, and
  Language Processing, IEEE Transactions on}, vol.~16, no.~5, pp. 980--988,
  2008.

\bibitem{stafylakis2012plda}
T.~Stafylakis, P.~Kenny, M.~Senoussaoui, and P.~Dumouchel, ``Plda using
  gaussian restricted boltzmann machines with application to speaker
  verification.'' in \emph{INTERSPEECH}, 2012.

\bibitem{salakhutdinov2009learning}
R.~Salakhutdinov, ``Learning deep generative models,'' Ph.D. dissertation,
  University of Toronto, 2009.

\bibitem{hinton2010practical}
G.~Hinton, ``A practical guide to training restricted boltzmann machines,''
  \emph{Momentum}, vol.~9, no.~1, p. 926, 2010.

\bibitem{wang2014gaussian}
N.~Wang, J.~Melchior, and L.~Wiskott, ``Gaussian-binary restricted boltzmann
  machines on modeling natural image statistics,'' \emph{arXiv preprint
  arXiv:1401.5900}, 2014.

\bibitem{cho2011improved}
K.~Cho, A.~Ilin, and T.~Raiko, ``Improved learning of gaussian-bernoulli
  restricted boltzmann machines,'' in \emph{Artificial Neural Networks and
  Machine Learning--ICANN 2011}.\hskip 1em plus 0.5em minus 0.4em\relax
  Springer, 2011, pp. 10--17.

\bibitem{hinton2002training}
G.~Hinton, ``Training products of experts by minimizing contrastive
  divergence,'' \emph{Neural computation}, vol.~14, no.~8, pp. 1771--1800,
  2002.

\bibitem{burda2014accurate}
Y.~Burda, R.~B. Grosse, and R.~Salakhutdinov, ``Accurate and conservative
  estimates of mrf log-likelihood using reverse annealing,'' \emph{arXiv
  preprint arXiv:1412.8566}, 2014.

\bibitem{senoussaoui2010vector}
M.~Senoussaoui, P.~Kenny, N.~Dehak, and P.~Dumouchel, ``An i-vector extractor
  suitable for speaker recognition with both microphone and telephone speech.''
  in \emph{Odyssey}, 2010, p.~6.

\bibitem{larcher2013phonetically}
A.~Larcher, K.~A. Lee, B.~Ma, and H.~Li, ``Phonetically-constrained plda
  modeling for text-dependent speaker verification with multiple short
  utterances,'' in \emph{Acoustics, Speech and Signal Processing (ICASSP), 2013
  IEEE International Conference on}.\hskip 1em plus 0.5em minus 0.4em\relax
  IEEE, 2013, pp. 7673--7677.

\bibitem{rajan2014single}
P.~Rajan, A.~Afanasyev, V.~Hautam{\"a}ki, and T.~Kinnunen, ``From single to
  multiple enrollment i-vectors: Practical plda scoring variants for speaker
  verification,'' \emph{Digital Signal Processing}, vol.~31, pp. 93--101, 2014.

\bibitem{khoury2014hierarchical}
E.~Khoury, L.~El~Shafey, M.~Ferras, and S.~Marcel, ``Hierarchical speaker
  clustering methods for the nist i-vector challenge,'' in \emph{Odyssey: The
  Speaker and Language Recognition Workshop}, no. EPFL-CONF-198439, 2014.

\bibitem{novoselov2014stc}
S.~Novoselov, T.~Pekhovsky, and K.~Simonchik, ``Stc speaker recognition system
  for the nist i-vector challenge,'' in \emph{Odyssey: The Speaker and Language
  Recognition Workshop}, 2014, pp. 231--240.

\bibitem{garcia2011analysis}
D.~Garcia-Romero and C.~Y. Espy-Wilson, ``Analysis of i-vector length
  normalization in speaker recognition systems.'' in \emph{Interspeech}, 2011,
  pp. 249--252.

\bibitem{brummer2007fusion}
N.~Brummer, L.~Burget, J.~H. Cernocky, O.~Glembek, F.~Grezl, M.~Karafiat, D.~A.
  Van~Leeuwen, P.~Matejka, P.~Schwarz, and A.~Strasheim, ``Fusion of
  heterogeneous speaker recognition systems in the stbu submission for the nist
  speaker recognition evaluation 2006,'' \emph{Audio, Speech, and Language
  Processing, IEEE Transactions on}, vol.~15, no.~7, pp. 2072--2084, 2007.

\bibitem{van2013distribution}
D.~A. van Leeuwen and N.~Br{\"u}mmer, ``The distribution of calibrated
  likelihood-ratios in speaker recognition,'' \emph{arXiv preprint
  arXiv:1304.1199}, 2013.

\end{thebibliography}

\end{document}